# End-to-end Phase Field Model Discovery Combining Experimentation, Crowdsourcing, Simulation and Learning

Md Nasim[1], Anter El-Azab[2], Xinghang Zhang[2], Yexiang Xue[1]
[1]Department of Computer Science, Purdue University, West Lafayette, IN, USA
[2]School of Materials Engineering, Purdue University, West Lafayette, IN, USA.
{mnasim, aelazab, xzhang98, yexiang}@purdue.edu

## Abstract

The availability of tera-byte scale experiment data calls for AI driven approaches which automatically discover scientific models from data. Nonetheless, significant challenges present in AI-driven scientific discovery: (i) The annotation of large scale datasets requires fundamental re-thinking in developing scalable crowdsourcing tools. (ii) The learning of scientific models from data calls for innovations beyond black-box neural nets. (iii) Novel visualization & diagnosis tools are needed for the collaboration of experimental and theoretical physicists, and computer scientists. We present PHASE-FIELD-LAB platform for end-to-end phase field model discovery, which automatically discovers phase field physics models from experiment data, integrating experimentation, crowdsourcing, simulation and learning. PHASE-FIELD-LAB combines (i) a streamlined annotation tool which reduces the annotation time (by $\approx 50 - 75\%$), while increasing annotation accuracy compared to baseline; (ii) an end-to-end neural model which automatically learns phase field models from data by embedding phase field simulation and existing domain knowledge into learning; and (iii) novel interfaces and visualizations to integrate our platform into the scientific discovery cycle of domain scientists. Our platform is deployed in the analysis of nano-structure evolution in materials under extreme conditions (high temperature and irradiation). Our approach reveals new properties of nano-void defects, which otherwise cannot be detected via manual analysis.

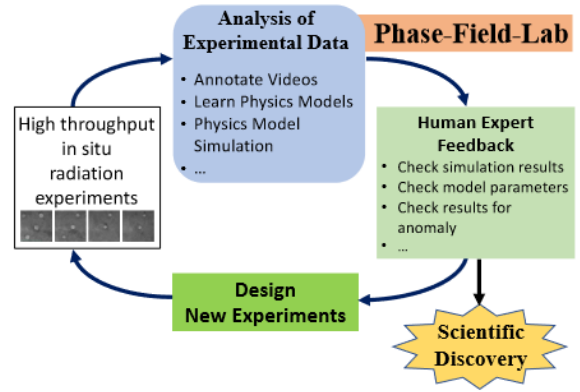

Figure 1: Scientific discovery workflow in material science domain, assisted by our PHASE-FIELD-LAB framework. PHASE-FIELD-LAB provides material scientists an integrated framework for data annotation, physics model learning and simulation-visualization of physics models.

# Introduction

Efficient methods to learn physics models from data have the potential to greatly accelerate scientific discovery and broaden our understanding of the physical world. The learning of physics models from data is a continuous process, which starts with data collection through carefully designed laboratory experiments or observation of natural phenomena. The amount of data collected for this pipeline of related analysis tasks can be overwhelming. For example, in the application domains considered in this paper, in situ radiation experiments to study material properties under extreme conditions generate high resolution high frame rate videos in the scale of terabytes Niu et al. [2020]. To automate the discovery of new scientific knowledge from these large scale datasets, scalable and reliable AI-driven data analysis methods are essential.

Existing AI-driven data analysis methods are often not readily applicable in the scientific discovery pipeline for a few reasons. First, a huge amount of labeled data is required to train the state of the art deep learning and other supervised machine learning methods. Such labeled data are often hard to obtain, since they require huge manual effort coming from domain experts and crowdworkers Fredriksson et al. [2020], Sheng and Zhang [2019], Weld et al. [2011]. Different applications/experiments may have unique characteristics, which makes it very difficult to use labeled data from one particular application for other applications Zhuang et al. [2020]. Second, existing black box neural network models often do not yield interpretable models Chakraborty et al. [2017], Vinuesa and Sirmacek [2021], Li et al. [2022] and may violate prior known physics principles. Such black box approach are not helpful for physicists trying to draw new insights from experimental data. Third, existing evaluation metrics for learned models such as accuracy, F1-score etc. are not sufficient for many applications, since models having similar scores according to these metrics can yield significantly different system dynamics. Qualitative methods such as simulation and visualization, on the contrary, are often preferred because they offer a straightforward interface to the domain experts Chatzimparmpas et al. [2020]. However, the development of these simulation and visualization techniques requires insights from domain experts as well as computational thinking from computer scientists.

In this paper, we address the problem of designing efficient integrated AI-driven scientific discovery framework, for discovering new physics models, in particular phase field models Kim et al. [1999], Millett et al. [2011], Kuhn and Müller [2010], Wheeler et al. [1992], Suzuki et al. [2002], directly from experimental data. Phase field models are widely used for solving a variety of interfacial problems. Here, we focus on the phase field model of *nano* size void defect evolution in materials Millett et al. [2011]. Learning such model of material defect evolution is essential for designing sustainable materials for extreme environments with high temperature and irradiation such as that inside a nuclear reactor. The analysis of the experimental data for this purpose requires streamlined annotation tools to collect high quality annotations from domain experts and crowdworkers in an efficient way, machine learning methods that take real world observation and domain expert inputs to learn practical physics models, and finally simulation and visualization tools for evaluation.

We introduce PHASE-FIELD-LAB – an integrated platform for end-to-end learning of phase field models from experiment data. PHASE-FIELD-LAB presents the material science researchers an efficient AI-based annotation tool that can be used to pixelwise annotate material defects in high dimensional video frames within minutes (i.e., $1000 \times 1000$ pixel video frame takes $< 6$ minutes). These annotated video frames can be used for automatically learning corresponding phase field models of material defect evolution. Additionally, researchers can also simulate and visualize outputs from the learned phase field model in our PHASE-FIELD-LAB platform. The scientific discovery loop assisted by our PHASE-FIELD-LAB is shown in Figure 1.

The annotation tool integrated in our PHASE-FIELD-LAB framework uses image segmentation Achanta et al. [2012] in the background and interactive user interface in the foreground for easy annotation. The automatic learning of phase field models is based on our novel partial differential equation (PDE) learning algorithm, where we use domain expert given constraints, simulation and ground truth annotations to learn phase field model parameters from video data. The simulation and visualization module is based on numerical method to solve partial differential equation in phase field models.

On evaluation, we find that our annotation module greatly reduces annotation time (by ≈ 50 − 75%) and provides more accurate annotations compared to baseline, and our learning module yields practical physics models that does not violate prior known physics constraints. We used VGG Image Annotation (VIA) Dutta et al. [2016] as the baseline for evaluating our annotation module. VIA was previously used for annotating defects in advanced STEM images of steels Roberts et al. [2019], thus acts as a suitable baseline for comparison (details of the user study including user demographics, expertise of the participants, test settings etc. are given later). The learned physics models from our learning module always satisfies prior known physics constraints, while competing baseline often violates the physical constraints on the possible parameter values. Compared to the baseline method, our learning module thus provides the material scientists with more practical physics models.

PHASE-FIELD-LAB is currently being used by material scientists for analyzing material defects evolution in metallic materials, such as the void defects shown in Figure 2. Using our tool provides the researchers the ability to scale up the analysis process Niu et al. [2020], which in turn has led to the discovery of new physics phenomena of the size fluctuations of the void defects in metal (Cu) under irradiation and high temperature Nasim et al. [2023].

Our key contribution in this work is the development of PHASE-FIELD-LAB , which provides an integrated platform combining modules for annotation, learning and simulation, for discovering phase field physics models from experimental data. Our PHASE-FIELD-LAB framework has already led to real-world scientific discovery of an interesting material defect size fluctuation property, and is now being used to accelerate scientific discovery in the domain of designing new materials for a sustainable energy future.

## Background : Phase field model

Phase field model is a widely used mathematical model for studying microstructure evolution and interfacial problems such as material defect evolution in high temperature and irradiation environment, grain growth, alloy decomposition, fracture mechanics, sintering, snowflake growth, collective cell migration etc. Millett et al. [2011], Fan and Chen [1997], Cahn [1961], Palmieri et al. [2015].

In phase field modeling, the system is described with a set of phase field variables. For our application domain in PHASE-FIELD-LAB – the analysis of nano-void defects in irradiated materials, the system is described with 3 phase field variables – $c_v, c_i$ and $\eta$ Millett et al. [2011]. The dynamics of these phase fields are governed by the Cahn-Hilliard equation Cahn and Hilliard [1958] for conserved phase fields, and Allen-Cahn equation Allen and Cahn [1972] for non-conserved phase fields:

$$\frac{\partial u}{\partial t} = \nabla \cdot (M \nabla \frac{\delta F}{\delta u}), \qquad \frac{\partial \eta}{\partial t} = -L \frac{\delta F}{\delta \eta}. \qquad (1)$$

Here, $u$ and $\eta$ represent conserved (i.e., $c_v, c_i$) and non-conserved phase fields respectively, $F$ is the system's free energy, $M$ is the diffusivity, $L$ is the mobility coefficient. $\nabla$ is first order spatial derivative. $M, L$, and the scalar parameters in $F$ represents different attributes of the system and the associated material species. These parameters are of great interest to physicists as they greatly affect material properties, and determine the long term system dynamics. For details of phase field model, we refer to the original text Millett et al. [2011].

## PHASE-FIELD-LAB : AI Platform for Physics Model Discovery

Our PHASE-FIELD-LAB framework is inspired by the need to efficiently learn the phase field model parameters of nano-void defect evolution in materials, although the general strategy used in the framework can be used to learn other types of phase field models, and more broadly other partial differential equation physics models.

Materials exposed to extreme conditions such as high temperature and irradiation suffer degradation over time due to the generation of different types of defects. In situ radiation experiment videos provide us with real-time observation of these defect generation and evolution under extreme conditions. The huge amount of experimental data generated during these experiments calls for automated methods to assist the material scientists in the analysis process.

PHASE-FIELD-LAB provides material scientists with an integrated platform to annotate video data, learn phase field models and evaluate the learned models with simulation-visualization, all in a single place.

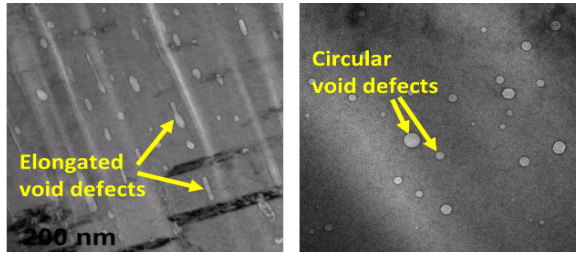

(a) Void defects of different shapes in irradiated Cu material

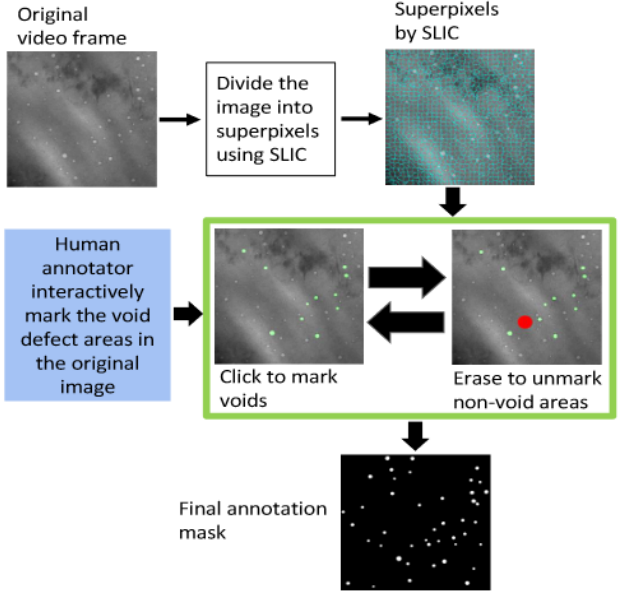

(b) Workflow of the annotation module in PHASE-FIELD-LAB

Figure 2: Nano-void defects in irradiated materials appear in different shapes, and our PHASE-FIELD-LAB annotation module can help annotate these defects very efficiently with minimal human effort. (a) **Left:** elongated void defects in irradiated Cu (112), **Right:** circular void defects in irradiated Cu (110), both captured by Transmission Electron Microscopy (TEM) imaging at nano-meter scale. (b) Workflow of our annotation module for labeling nano-void defects in video frames. Our annotation module first segments the video frame image into superpixels, and then human annotators can mark these superpixels through an interactive interface by clicking and erasing.

As shown in Figure 1, PHASE-FIELD-LAB consists of multiple modules:

1. **Annotation Tool.** The AI-powered annotation tool provides human users a quick and efficient interactive interface to pixelwise annotate high resolution in situ videos.

2. **Physics Learning with Human Expert in the Loop.** The learning module provides an easy interface to material scientists for automatic learning of physics models and material properties, using domain expertise and annotations from the annotation module.

3. **Simulation and Visualization.** With our simulation-visualization module, users can simulate system dynamics under a phase field physics model and visualize the dynamics, thus qualitatively evaluate the learned physics models, and build intuition about model parameters' role in nano-structure dynamics.

## AI-based Annotation Tool for Video Frames

The easy and scalable annotation of video data saves human effort and also facilitates the application of supervised learning methods such as deep learning. Classic image processing techniques i.e. filtering, thresholding etc. do not work well with variable illumination and heavy noise, and our annotation module is designed to be robust against such factors. The general workflow with our annotation tool is shown in Figure 2, which works by first segmenting the images into superpixels, and then facilitating human input through an interactive user interface.

**Superpixel Clustering.** Our annotation tool takes as input an image showing nano-void defects, and divides the image into *superpixels* by combining nearby pixels having similar appearance. We use the Simple Linear Iterative Clustering (SLIC) algorithm Achanta et al. [2012] to generate these superpixels. The SLIC algorithm takes as input an image $I$ and the number of superpixels $K$, and then segments the image $I$ into $K$ superpixels as output. We manually set the value of $K$ much higher than the number of nano-voids in the video. In this way we generate redundant superpixels, to counter the slight variation in appearance within the same nano-void due to illumination, noise etc.

**Interactive User Interface.** After this over-segmentation by superpixel clustering is completed in the background, we provide an interactive interface to human annotators, showing the original image containing nano-void defects. The users can now mark and dif-ferentiate the void defects from background, by mouse-clicking on the voids in the image. As the users interact by clicking, the superpixel corresponding to the user-click is marked as part of a void defect. In case a superpixel contains both void defect and background region together, the users can unmark the background portion of the superpixel, by activating the eraser mode in the tool, and then using mouse-click-and-drag on background region. In this way, a human annotator can pixelwise label all the void defects in the whole image, using just a few clicks and erase within minutes.

**Phase Field Model Learning Module**

**Problem Description.** Suppose we have a series of noisy in situ video frames $\mathcal{V} = \{v_1, v_2, \ldots, v_T\}$, and pixelwise annotations $\mathcal{A}$ of a subset of these frames. The annotations $\mathcal{A}$ marks the void defects pixelwise in these video frames. We also have a set of constraints on the possible phase field model parameter values, a range of possible values $[\theta_{min}, \theta_{max}]$ provided by human domain experts. We want to annotate void defects in all the video frames and also learn the phase field model parameters $\theta$ in the range $[\theta_{min}, \theta_{max}]$, that best fits the defect evolution dynamics in the video.

**Phase Field Model Learning with Human Input.** Given in situ video frames $\mathcal{V}$ and partial annotations $\mathcal{A}$ obtained from our annotation module, we can automatically annotate the entire video and also learn the phase filed model by using a combination of two neural networks - one to recognize phase field variables from video frames (Recognition net) and another to learn the evolution dynamics via simulation (Neural Differential Equation Net). Such 2-Network architecture is named Neuradiff and was first proposed in Xue et al. [2021]. The triage loss function in Neuradiff penalizes the difference between neural network output, simulation results and human annotations. However, while Neuradiff embeds PDEs of phase field model into learning, domain expert input is not incorporated in the learning approach. As a result, the learned model parameters from Neuradiff can violate known physical constraints. We extend the Neuradiff loss function by adding penalty terms to the loss function for violating domain expert given constraints.

Our learning module works as follows: suppose we have two video frames $v_o, v_T$ at time $t = 0$ and $t = T$ respectively, and associated annotations $a_0, a_T$. We use the Recognition neural network to extract the phase field variable $u_0$ from $v_0$. We also use the same Recognition net to extract phase field $u_{T(rec)}$ from $v_T$. Using the Neural Differential Equation Net, we simulate the evolution of $u_0$ for a time period $T$ to obtain $u_{T(sim)}$. The PDE physics model is embedded in the Neural Differential Equation Net, and this network has the same parameters $\theta$ as the PDE physics model. If both the Recognition net and Neural Differential Equation Net are trained perfectly, then $a_T, u_{T(rec)}$ and $u_{T(sim)}$ should all match. We add regularization terms for violating the expert given range $[\theta_{min}, \theta_{max}]$ on possible parameter values. Adding all these up, our final loss function is:

$$\min_{\theta} \mathcal{L}_{mismatch}(a_T, u_{T(rec)}, u_{T(sim)}) \\ + \lambda_1 \max\{0, \theta_{min} - \theta\} + \lambda_2 \max\{0, \theta - \theta_{max}\}) \quad (2)$$

Here, $\lambda_1, \lambda_2$ are hyperparameters. We then optimize for $\theta$ by backpropagating error gradients via stochastic gradient descent.

**Simulation and Visualization Module**

To visualize and evaluate learned phase field physics models, we simulate the system dynamics by solving the PDEs of phase field model (Equation 1), and then use Python to render the simulation outputs into a video. Solving the PDEs in essence means solving an initial value problem, where given an initial system state $u_0$ at time $t = 0$, and a PDE of the general format $\frac{\partial u}{\partial t} = f(u, x, t, \theta)$, we solve for system state $u_T$ at time $t = T$. Here, $x$ represents spatial coordinates, $t$ represents time, $\theta$ denotes scalar model parameters.

To solve PDEs, we use finite difference method, which approximate partial derivatives with finite quotients i.e., $\frac{\partial u}{\partial t}$ can be approximated as $\frac{\partial u}{\partial t} \approx \frac{u_{t+1} - u_t}{\Delta t}$, and then use forward Euler time marching to obtain the final solution. An example of simulation output from this module is shown in Figure 4.

## System Deployment

The 3 different modules of our PHASE-FIELD-LAB framework were originally developed using the Python programming language, OpenCV, scikit-image, PyTorch and torchvision. We deployed a simple web-based interface using HTML, CSS and jquery for

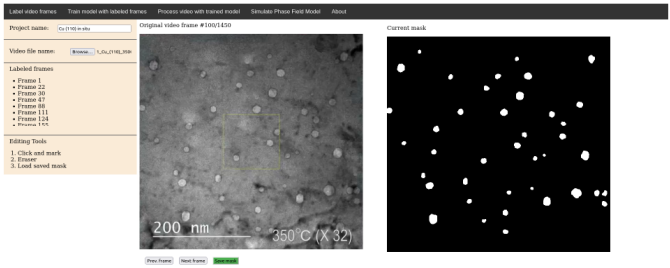

(a) Annotation module for video frame images

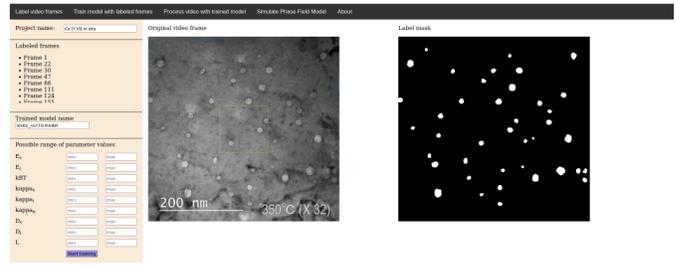

(b) Module for learning phase field model from annotated video data

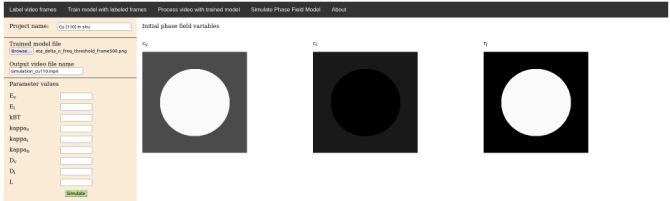

(c) Simulation module for phase field model of void defect evolution

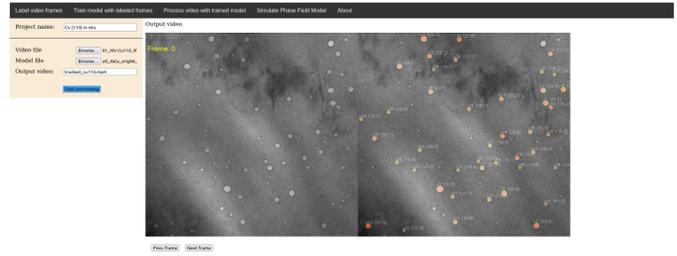

(d) Annotate entire videos with learned model

Figure 3: Our PHASE-FIELD-LAB gives material scientists the option to (a) label a few video frames, (b) use the labeled frames to train a supervised machine learning model to learn underlying physics, (c) use simulation to qualitatively assess learned physics models and (d) apply learned machine learning model to annotate material defects seen in entire video.

all 3 modules in PHASE-FIELD-LAB framework: annotation, simulation with visualization, and learning physics models for easy use by material scientists. The web interface is shown in Figure 3.

For the annotation tool, users can select the video file, select any frame in the video and then annotate. For learning phase field model parameters, users can select the annotated video frames, and also specify the range of possible values for the model, before starting the learning iterations. In the simulation module, users can set the model parameter values, simulate void evolution dynamics and save the output as a video. Users can also annotate all void defects in a video by selecting a model from available trained models.

## PHASE-FIELD-LAB Evaluation

We conducted experiments to evaluate the annotation module and learning module in PHASE-FIELD-LAB .
**Evaluation of Annotation Module.** To evaluate the annotation module, we conducted a user study with 10 volunteers. The volunteers were all male graduate students, in age range 24-32 years, average age of 27 years. We only included participants who were unfamiliar with both our annotation module and baseline VIA Dutta et al. [2016]. For each participant, we conducted a separate training session, with a small demonstration on how to use the two different annotation tools – our PHASE-FIELD-LAB annotation module and VIA.

**Annotation Task.** We asked each participant to annotate void defects in 2 different in situ experiment video frame, using both our annotation module and VIA. Both video frame images had $1024 \times 1024$ pixel resolution, one containing approximately 80 visible circular nano-void defects, while the other containing a mixture of circular and elongated void defects ($\approx 30$ circular, $\approx 40$ elongated). We used the same 2 images for all 10 participants.

**Evaluation Criteria.** We measured 2 metrics to evaluate our annotation module – 1) time required for annotation, and 2) intersection-over-union (IOU), also known as Jaccard similarity coefficient, which measures the similarity between annotations and ground truth. For all metrics, we also computed unbiased effect size Cohen's $d$ Goulet-Pelletier and Cousineau [2018] for paired samples, corrected for overestimation error due to small sample size, to see if the results have practical significance.

Table 1: Annotation time and accuracy for labeling nano-void defects in experimental videos

| Void defects shape in video frames | Annotation tool | Average annotation time (s) per video frame ↓ | Average IOU (%) of annotation and ground truth ↑ |
|---|---|---|---|
| Circular | PHASE-FIELD-LAB (Ours) | **282 ± 38** | **75.2±2.82** |
|  | VIA (Baseline) | 558 ± 60 | 53.8±2.86 |
| Elongated, circular | PHASE-FIELD-LAB (Ours) | **411 ± 47** | **74.4±3.78** |
|  | VIA (Baseline) | 1637 ± 202 | 39.1±7.39 |

**Results.** The experimental results are summarized in Table 1. Overall, we found that our annotation tool was able to reduce the annotation times for both circular voids (by $49.84\%$ average per frame, Cohen's $d = 5.48$), and mix of elongated and circular voids (by $74.7\%$ average per frame, Cohen's $d = 8.36$) compared to the baseline VIA tool. The Cohen's $d$ value of 5.48 for circular void annotation times implies that, the mean annotation times for PHASE-FIELD-LAB and VIA differed by 5.48 times the average standard deviation. Annotations from PHASE-FIELD-LAB had higher IOU compared to baseline VIA (for circular voids Cohen's $d = 7.53$, and for mixed voids Cohen's $d = 6.02$). Large effect size Cohen's $d$ values in all the metrics implies that our annotation module has significant improvement over baseline for all computed metrics.

On average, it took the users less time to annotate circular voids compared to the elongated ones, due to increased use of erasure feature in ours, and drawing complex polygon shape in VIA. Performance of both annotation tools suffered in presence of noise in video frames. In post experiment survey, all our participants commented that they preferred our annotation module to the baseline VIA, mentioning that click and erase approach was "much easier" and "fun".

**Evaluation of Learning Module.**

To evaluate the performance of the learned phase field model from in situ experiment data, we synthesized a $128 \times 128$ resolution video, depicting the evolution of 2 voids of different sizes, using the phase field model in Millett et al. [2011]. We then used this video and associated partial annotations to automatically annotate all video frames and learn the phase field model of the void evolution shown in the video. For training both baseline vanilla Neuradiff Xue et al. [2021], and our learning module, we used 1000 randomly chosen video frames for training, another 100 for hyperparameter tuning and another 300 for testing. Additionally, for our learning module, we specified the possible ranges of 9 out of 14 scalar parameters.

**Results.** We tested the two learning approaches - ours and baseline on two objectives: 1) annotate video frames and 2) predict void defect dynamics. Our learning module performs similarly as the baseline vanilla Neuradiff in annotating the unseen video frames (both $\approx 96\%$ pixelwise accuracy). The learned phase field models from both baseline and our learning module produce similar evolution dynamics, and have similar mean squared error (MSE) after 100 timestep simulation (both $\approx 3.3 \times 10^{-4}$ per pixel). The tracking and simulation results from PHASE-FIELD-LAB are in Figure 4.

Upon looking into the actual parameter values, we found that baseline Neuradiff trained models violate at least one out of the 9 "known range" constraints on each of multiple runs. In contrast, our trained models' parameters always satisfy the range constraints. It is important to note that the baseline trained models violating known physics priors are of no practical use in real world practical scenarios. Our learning module thus yields more practical physics models.

## Impact of PHASE-FIELD-LAB in Scientific Discovery

**Scalable Analysis of in situ Experiment Video Data.** Our PHASE-FIELD-LAB has been deployed in the real

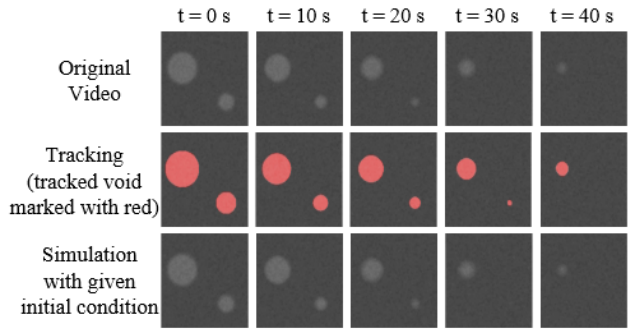

Figure 4: Our PHASE-FIELD-LAB provides high accuracy tracking of nano-void defects from video (second row), and phase field model which yields similar dynamics as ground truth video data (third row).

world to efficiently analyze high volume the experimental in situ video data Nasim et al. [2023], Niu et al. [2020, 2021]. Manual annotation of a single 10-minute in situ video can take upto 3.75 months Xue et al. [2021]. Our PHASE-FIELD-LAB can accomplish this in less than a day, combining AI-driven automation and domain expert supervision, thus yielding an accelerated scientific discovery workflow.

**Discovery of Nano-void Defect Size Fluctuation under Heavy Ion Irradiation.** Detail analysis of in situ video data with our PHASE-FIELD-LAB has led to the discovery of void defect size fluctuation in metallic materials under high temperature and irradiation Nasim et al. [2023]. Previously, partial analysis of the in situ video data by manual efforts revealed that the void defects size decreases monotonically during irradiation Fan et al. [2019], Chen et al. [2015]. Our analysis revealed that the defect size change is not perfectly monotonic, rather mimics a random walk, fluctuating over time as shown in Figure 5.

Using the annotation module of our PHASE-FIELD-LAB framework, 1% of the video frames in an 8-minute in situ video was annotated by human experts. These annotations were then used to train a U-Net model and annotate defects in the entire video. Manual measurements of void defects size and additional in situ experiments were carried out to confirm size fluctuation observation. Phase field model simulation was performed to analyze the cause behind such size fluctuation. All additional experiments and analysis confirmed the void defect size fluctuation phenomenon. Thus, our PHASE-FIELD-LAB framework played a critical role in a real world scientific discovery in material science domain.

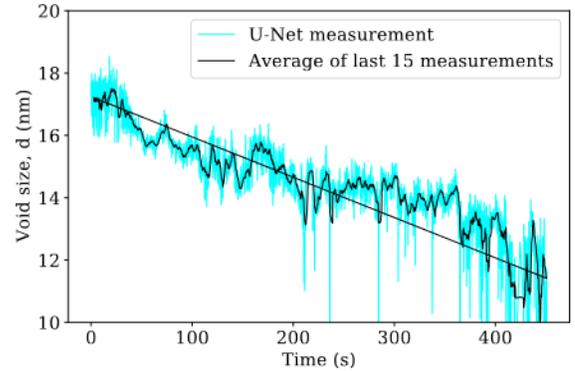

Figure 5: Our PHASE-FIELD-LAB framework helped discover the size fluctuation of nano-void defects in metallic materials under irradiation Nasim et al. [2023]. Here, we see the size fluctuation of a single nano-void defect over time, during Kr irradiation of Cu (110) at $350\ °C$. Void defect size fluctuation was confirmed by manual measurements, and the possible cause behind such fluctuation was investigated through phase field simulations.

## Conclusion

In this paper, we present PHASE-FIELD-LAB – an integrated platform for annotating video frames, learning the phase field physics model of nano-void defect evolution directly from video data, and performing simulation and visualization with the learned models. Our framework addresses the fundamental challenges in learning physics models from data - the lack of easy annotation process, the lack of expert guidance in the automatic learning of physics models from data, and the lack of integrated tools for experts to qualitatively assess a particular hypothesis/physics model. Our model led to the discovery of nano-void size fluctuation in materials under high temperature and irradiation. The general approach used in our framework can be easily adapted to other phase field models, and broadly other partial differential equation models as well. Thus our framework has the potential to greatly accelerate the scientific discovery in diverse domains.

## Acknowledgements

This research was supported by NSF grants CCF-1918327.

# References

T. Niu, M. Nasim, R. Annadanam, C. Fan, Jin Li, Zhongxia Shang, Y. Xue, Anter El-Azab, H. Wang, and X. Zhang. Recent studies on void shrinkage in metallic materials subjected to in situ heavy ion irradiations. *JOM*, 72, 09 2020. doi: 10.1007/s11837-020-04358-3.

Teodor Fredriksson, David Issa Mattos, Jan Bosch, and Helena Holmström Olsson. Data labeling: An empirical investigation into industrial challenges and mitigation strategies. In Maurizio Morisio, Marco Torchiano, and Andreas Jedlitschka, editors, *Product-Focused Software Process Improvement*, pages 202–216, Cham, 2020. Springer International Publishing. ISBN 978-3-030-64148-1.

Victor S Sheng and Jing Zhang. Machine learning with crowdsourcing: A brief summary of the past research and future directions. In *Proceedings of the AAAI conference on artificial intelligence*, volume 33 (01), pages 9837–9843, 2019.

Daniel S Weld, Peng Dai, et al. Human intelligence needs artificial intelligence. In *Workshops at the Twenty-Fifth AAAI Conference on Artificial Intelligence*, 2011.

Fuzhen Zhuang, Zhiyuan Qi, Keyu Duan, Dongbo Xi, Yongchun Zhu, Hengshu Zhu, Hui Xiong, and Qing He. A comprehensive survey on transfer learning. *Proceedings of the IEEE*, 109(1):43–76, 2020.

Supriyo Chakraborty, Richard Tomsett, Ramya Raghavendra, Daniel Harborne, Moustafa Alzantot, Federico Cerutti, Mani Srivastava, Alun Preece, Simon Julier, Raghuveer M Rao, et al. Interpretability of deep learning models: A survey of results. In *2017 IEEE smartworld, ubiquitous intelligence & computing, advanced & trusted computed, scalable computing & communications, cloud & big data computing, Internet of people and smart city innovation (smartworld/SCALCOM/UIC/ATC/CBDcom/IOP/SCI)*, pages 1–6, San Francisco, CA, USA, 2017. IEEE. doi: 10.1109/UIC-ATC.2017.8397411.

Ricardo Vinuesa and Beril Sirmacek. Interpretable deep-learning models to help achieve the sustainable development goals. *Nature Machine Intelligence*, 3(11):926–926, 2021.

Xuhong Li, Haoyi Xiong, Xingjian Li, Xuanyu Wu, Xiao Zhang, Ji Liu, Jiang Bian, and Dejing Dou. Interpretable deep learning: Interpretation, interpretability, trustworthiness, and beyond. *Knowledge and Information Systems*, 64(12):3197–3234, 2022.

Angelos Chatzimparmpas, Rafael M Martins, Ilir Jusufi, and Andreas Kerren. A survey of surveys on the use of visualization for interpreting machine learning models. *Information Visualization*, 19(3): 207–233, 2020.

Seong Gyoon Kim, Won Tae Kim, and Toshio Suzuki. Phase-field model for binary alloys. *Physical review e*, 60(6):7186, 1999.

Paul C Millett, Anter El-Azab, Srujan Rokkam, Michael Tonks, and Dieter Wolf. Phase-field simulation of irradiated metals: Part i: Void kinetics. *Computational materials science*, 50(3):949–959, 2011.

Charlotte Kuhn and Ralf Müller. A continuum phase field model for fracture. *Engineering Fracture Mechanics*, 77(18):3625–3634, 2010.

Adam A Wheeler, William J Boettinger, and Geoffrey B McFadden. Phase-field model for isothermal phase transitions in binary alloys. *Physical Review A*, 45(10):7424, 1992.

Toshio Suzuki, Machiko Ode, Seong Gyoon Kim, and Won Tae Kim. Phase-field model of dendritic growth. *Journal of Crystal Growth*, 237:125–131, 2002.

Radhakrishna Achanta, Appu Shaji, Kevin Smith, Aurelien Lucchi, Pascal Fua, and Sabine Süsstrunk. Slic superpixels compared to state-of-the-art superpixel methods. *IEEE transactions on pattern analysis and machine intelligence*, 34(11):2274–2282, 2012.

A. Dutta, A. Gupta, and A. Zissermann. VGG image annotator (VIA). http://www.robots.ox.ac.uk/ vgg/software/via/, 2016.

Graham Roberts, Simon Y Haile, Rajat Sainju, Danny J Edwards, Brian Hutchinson, and Yuanyuan Zhu.

Deep learning for semantic segmentation of defects in advanced stem images of steels. *Scientific reports*, 9(1):1–12, 2019.

M. Nasim, Sreekar Rayaprolu, T. Niu, C. Fan, Z. Shang, Jin Li, H. Wang, A. El-Azab, Y. Xue, and X. Zhang. Unraveling the size fluctuation and shrinkage of nanovoids during in situ radiation of cu by automatic pattern recognition and phase field simulation. *Journal of Nuclear Materials*, 574:154189, 2023. ISSN 0022-3115. doi: https://doi.org/10.1016/j.jnucmat.2022.154189. URL https://www.sciencedirect.com/science/article/pii/S0022311522006687.

Danan Fan and L-Q Chen. Computer simulation of grain growth using a continuum field model. *Acta Materialia*, 45(2):611–622, 1997.

John W Cahn. On spinodal decomposition. *Acta metallurgica*, 9(9):795–801, 1961.

Benoit Palmieri, Yony Bresler, Denis Wirtz, and Martin Grant. Multiple scale model for cell migration in monolayers: Elastic mismatch between cells enhances motility. *Scientific reports*, 5(1):11745, 2015.

John W Cahn and John E Hilliard. Free energy of a nonuniform system. i. interfacial free energy. *The Journal of chemical physics*, 28(2):258–267, 1958.

Samuel Miller Allen and John W Cahn. Ground state structures in ordered binary alloys with second neighbor interactions. *Acta Metallurgica*, 20(3):423–433, 1972.

Yexiang Xue, Md Nasim, Maosen Zhang, Cuncai Fan, Xinghang Zhang, and Anter El-Azab. Physics knowledge discovery via neural differential equation embedding. In Yuxiao Dong, Nicolas Kourtellis, Barbara Hammer, and Jose A. Lozano, editors, *Machine Learning and Knowledge Discovery in Databases. Applied Data Science Track*, pages 118–134, Cham, 2021. Springer International Publishing. ISBN 978-3-030-86517-7.

Jean-Christophe Goulet-Pelletier and Denis Cousineau. A review of effect sizes and their confidence intervals, part i: The cohen'sd family. *The Quantitative Methods for Psychology*, 14(4):242–265, 2018.

Tongjun Niu, Yifan Zhang, Jaehun Cho, Jin Li, Haiyan Wang, and Xinghang Zhang. Thermal stability of immiscible cu-ag/fe triphase multilayers with triple junctions. *Acta Materialia*, 208:116679, 2021.

C Fan, ARG Sreekar, Z Shang, Jin Li, M Li, H Wang, A El-Azab, and X Zhang. Radiation induced nanovoid shrinkage in cu at room temperature: An in situ study. *Scripta Materialia*, 166:112–116, 2019.

Y Chen, K Y Yu, Y Liu, S Shao, H Wang, MA Kirk, J Wang, and X Zhang. Damage-tolerant nanotwinned metals with nanovoids under radiation environments. *Nature communications*, 6(1):7036, 2015.